\documentclass[pdflatex,sn-basic]{sn-jnl}

\usepackage{graphicx}
\usepackage{multirow}
\usepackage{amsmath,amssymb,amsfonts}
\usepackage{amsthm}
\usepackage{xcolor}
\usepackage{booktabs}
\usepackage{url}
\usepackage{placeins}

\newif\ifanonymous
\ifdefined\anonymoussubmission \anonymoustrue \else \anonymousfalse \fi
\newif\ifrepolink
\repolinkfalse
\newif\ifreproduction
\reproductiontrue

\newcommand{\AccDirCI}{[+0.55, +1.01]}
\newcommand{\AccDirDelta}{+0.78}
\newcommand{\AccDirP}{p<0.001}
\newcommand{\AccDirSignP}{p<0.001}
\newcommand{\AccDirWins}{14/15}
\newcommand{\AccDisCI}{[-0.66, +0.20]}
\newcommand{\AccDisDelta}{-0.23}
\newcommand{\AccDisP}{p=0.362}
\newcommand{\AccDisSignP}{p=0.607}
\newcommand{\AccDisWins}{6/15}
\newcommand{\AccPetsCI}{[+0.33, +0.71]}
\newcommand{\AccPetsDelta}{+0.52}
\newcommand{\AccPetsP}{p<0.001}
\newcommand{\AccPetsSignP}{p=0.002}
\newcommand{\AccPetsWins}{10/10}

\newcommand{\AlignDirFirst}{0.85}
\newcommand{\AlignDirLast}{0.28}
\newcommand{\AlignDisFirst}{0.66}
\newcommand{\AlignDisLast}{0.38}

\newcommand{\AlignPetsLast}{0.57}

\newcommand{\AlignPetsTrainLast}{0.95}
\newcommand{\AlignRegDiff}{0.0002}
\newcommand{\AlphaBonf}{0.0125}

\newcommand{\AlphaHalfAlignLast}{0.29}

\newcommand{\AlphaOneAlignLast}{0.24}

\newcommand{\AlphaOneInfMean}{0.2}
\newcommand{\AlphaOneInfUp}{36}

\newcommand{\AlphaTenAccTOSTP}{p<0.001}

\newcommand{\AlphaTenAlignFirst}{0.95}

\newcommand{\AlphaTenPairs}{50}

\newcommand{\AlphaTenSpread}{1.5}

\newcommand{\AlphaTenStepsMax}{78}
\newcommand{\AlphaTenStepsMin}{47}

\newcommand{\AlphaTenthAlignFirst}{0.85}
\newcommand{\AlphaTenthAlignLast}{0.53}

\newcommand{\AlphaTenthInfUp}{98}

\newcommand{\AlphaTenthPairs}{60}

\newcommand{\AlphaTenthSpread}{39.6}

\newcommand{\AlphaTenthStepsMax}{240}
\newcommand{\AlphaTenthStepsMin}{1}

\newcommand{\AlphaTrendAccP}{p=0.001}
\newcommand{\AlphaTrendAccRho}{-0.50}
\newcommand{\AlphaTrendMeanP}{p<0.001}
\newcommand{\AlphaTrendMeanRho}{-0.77}
\newcommand{\BatchCifar}{32}
\newcommand{\BatchPets}{64}
\newcommand{\BonfThreshold}{0.0024}
\newcommand{\CosAvgDir}{+0.005}
\newcommand{\CosAvgDis}{-0.151}
\newcommand{\CosAvgPets}{+0.051}

\newcommand{\CtrlDirBonf}{0.0125}

\newcommand{\CtrlDirMeanInf}{7.7}
\newcommand{\CtrlDirRot}{+0.17}

\newcommand{\CtrlDirRotHarm}{42}

\newcommand{\CtrlDirRotP}{p=0.337}
\newcommand{\CtrlDirRotRotation}{9.9}

\newcommand{\CtrlDirUnif}{+0.96}

\newcommand{\CtrlDirUnifHarm}{1.7}
\newcommand{\CtrlDirUnifInf}{79}

\newcommand{\CtrlDirUnifP}{p=0.003}

\newcommand{\CtrlDirUnifWins}{9/10}

\newcommand{\CtrlDirUpInf}{84}

\newcommand{\CtrlDirUpRotation}{16.2}

\newcommand{\CtrlDirUpVsRot}{+0.74}

\newcommand{\CtrlDirUpVsRotP}{p=0.003}

\newcommand{\CtrlDirUpVsRotWins}{9/10}
\newcommand{\CtrlDirUpVsUnif}{-0.04}

\newcommand{\CtrlDirUpVsUnifTOSTP}{p=0.009}

\newcommand{\CtrlDirWouldMean}{9.6}

\newcommand{\CtrlDirWouldUnif}{17.0}
\newcommand{\CtrlDirWouldUp}{16.2}

\newcommand{\CtrlPetsRot}{-0.01}

\newcommand{\CtrlPetsRotDev}{3.9\times10^{-4}}

\newcommand{\CtrlPetsRotP}{p=0.786}

\newcommand{\CtrlPetsUnif}{+1.37}

\newcommand{\CtrlPetsUnifLoss}{-27.6}

\newcommand{\CtrlPetsUnifP}{p<0.001}

\newcommand{\CtrlPetsUnifWins}{10/10}

\newcommand{\CtrlPetsUp}{+0.52}

\newcommand{\CtrlPetsUpLoss}{-12.7}

\newcommand{\CtrlPetsUpVsRot}{+0.53}

\newcommand{\CtrlPetsUpVsRotP}{p=0.001}

\newcommand{\CtrlPetsUpVsRotWins}{10/10}
\newcommand{\CtrlPetsUpVsUnif}{-0.84}

\newcommand{\CtrlPetsUpVsUnifP}{p=0.001}

\newcommand{\DLTenAccChange}{-0.46}

\newcommand{\DLTenAccChangeP}{p<0.001}
\newcommand{\DLTenAccDelta}{+0.07}
\newcommand{\DLTenAccP}{p=0.320}

\newcommand{\DLTenAccTOSTP}{p<0.001}
\newcommand{\DLTenAccWins}{6/10}
\newcommand{\DLTenGainMean}{+6.13}

\newcommand{\DLTenSteps}{2.8}

\newcommand{\DdDirMax}{0.41}
\newcommand{\DdDirMin}{0.07}

\newcommand{\DdDisMax}{0.16}
\newcommand{\DdDisMin}{0.02}

\newcommand{\DdPetsLast}{0.26}
\newcommand{\DdPetsMax}{0.79}

\newcommand{\DecileDirP}{p=0.038}

\newcommand{\DropShareDir}{0.8}
\newcommand{\DropSharePets}{20.7}

\newcommand{\DualProjDev}{5.3\times10^{-8}}
\newcommand{\DualProjN}{800}

\newcommand{\EqDisMeanCI}{[-0.38, +0.38]}

\newcommand{\EqDisMeanP}{p=0.018}

\newcommand{\FBAccDelta}{-7.97}
\newcommand{\FBAccP}{p=0.084}

\newcommand{\FBInfMain}{450}

\newcommand{\FBNMain}{450}

\newcommand{\FBStdDelta}{+2.29}
\newcommand{\FBStdP}{p=0.086}

\newcommand{\FBWorstDelta}{-5.76}
\newcommand{\FBWorstP}{p=0.007}

\newcommand{\FairCells}{12}
\newcommand{\FamilyBonf}{4}
\newcommand{\FamilySig}{8}
\newcommand{\FamilySize}{21}

\newcommand{\FloorAvgDis}{60}

\newcommand{\FloorFive}{-0.25}

\newcommand{\FloorPeakDis}{81}

\newcommand{\FloorTwenty}{-0.0526}
\newcommand{\GHAccDelta}{+0.69}

\newcommand{\GHInf}{48}

\newcommand{\GHN}{75}
\newcommand{\GHPmax}{0.907}
\newcommand{\GHPmin}{0.002}
\newcommand{\GHRounds}{25}

\newcommand{\InfDir}{750}
\newcommand{\InfDirPct}{83}
\newcommand{\InfDis}{57}
\newcommand{\InfDisPct}{6}

\newcommand{\KDir}{20}
\newcommand{\KDis}{5}
\newcommand{\KPets}{20}
\newcommand{\LadderDir}{29 \to 33 \to 34 \to 34}
\newcommand{\LadderDis}{24 \to 18 \to 18 \to 18}
\newcommand{\LadderFB}{9 \to 1 \to 0 \to 0}

\newcommand{\LadderNDir}{240}
\newcommand{\LadderNDis}{60}
\newcommand{\LadderNFB}{60}
\newcommand{\MaxHarmDir}{9}

\newcommand{\MomentumFactor}{530}
\newcommand{\NAll}{2200}
\newcommand{\NDir}{900}

\newcommand{\NPets}{400}
\newcommand{\NormErrPlain}{5.9\times10^{-3}}
\newcommand{\NormErrReg}{6.0\times10^{-7}}
\newcommand{\NormRatioMax}{3.02}
\newcommand{\NormRatioMin}{1.01}

\newcommand{\PPAcc}{+0.04}

\newcommand{\PPAccP}{p=0.823}

\newcommand{\PPAccTOSTP}{p=0.017}
\newcommand{\PPAccWins}{5/10}
\newcommand{\PPClients}{10}

\newcommand{\PPFullAcc}{+0.92}

\newcommand{\PPFullInf}{84}

\newcommand{\PPFullRot}{16.2}
\newcommand{\PPFullShift}{14.1}

\newcommand{\PPInf}{17}

\newcommand{\PPRot}{8.1}
\newcommand{\PPRoundSeeds}{600}

\newcommand{\PPShift}{7.7}

\newcommand{\PairsDir}{57}
\newcommand{\PairsDis}{99}
\newcommand{\PairsPets}{37}
\newcommand{\PlainCosDir}{0.216}
\newcommand{\PlainCosDis}{0.099}
\newcommand{\ProbeLastDir}{60}

\newcommand{\ProbeRoundsDir}{12}
\newcommand{\ProbeRoundsDis}{12}
\newcommand{\ProbedDir}{90}
\newcommand{\ProbedDis}{90}

\newcommand{\RiseMeanDirMean}{7.92}
\newcommand{\RiseMeanDirProj}{2.42}

\newcommand{\RiseRoundsDirProj}{8}

\newcommand{\RiseRoundsDisProj}{10}

\newcommand{\RotMeanPets}{6.44}
\newcommand{\RotMedMax}{17}
\newcommand{\RotMedMin}{6}
\newcommand{\RotShareDirCos}{35}

\newcommand{\RotShareDisNorms}{55}

\newcommand{\RotWorstDeg}{25.1}
\newcommand{\RoundsDir}{60}

\newcommand{\RoundsPets}{40}
\newcommand{\SeedsDir}{15}

\newcommand{\SeedsPets}{10}
\newcommand{\SignFlipsMax}{3}
\newcommand{\SizeSpreadMax}{16.6}
\newcommand{\SizeSpreadMin}{3.0}

\newcommand{\StepDirLastBetter}{15/15}

\newcommand{\StepDisFirstBetter}{2/15}

\newcommand{\StepPetsBetter}{46}

\newcommand{\StepPetsN}{50}

\newcommand{\StepShrinkFB}{64}
\newcommand{\StepsDir}{62}
\newcommand{\StepsDirMax}{165}
\newcommand{\StepsDirMin}{7}
\newcommand{\StepsDis}{250}

\newcommand{\StepsPets}{1.8}
\newcommand{\StepsPetsMax}{3}
\newcommand{\StepsPetsMin}{1}

\newcommand{\SymErrDir}{5.6}
\newcommand{\SymErrDis}{2.1}

\newcommand{\ThirdInputMin}{0.9986}
\newcommand{\TrajDir}{63}
\newcommand{\TrajDirAccFive}{-0.87}

\newcommand{\TrajDirAccFiveWins}{3/15}

\newcommand{\TrajDirPct}{7.0}

\newcommand{\TrajDirPosFrom}{18}
\newcommand{\TrajDis}{1}

\newcommand{\TrajPetsAccTen}{+0.68}

\newcommand{\TrajPetsAccTenWins}{10/10}

\newcommand{\TrajPetsAllAccTen}{+0.67}

\newcommand{\TrajPetsAllLossLast}{-6.7}

\newcommand{\TrajPetsLossLast}{-12.7}
\newcommand{\TrajPetsLossLastWins}{10/10}
\newcommand{\TrajPetsLossOne}{-0.6}
\newcommand{\TrajPetsLossOneP}{p=0.006}
\newcommand{\TrajPetsLossOneWins}{9/10}

\newcommand{\TrajPetsPeak}{+1.14}
\newcommand{\TrajPetsPeakRound}{19}

\newcommand{\UnrepDir}{44}
\newcommand{\UnrepDis}{72}

\newcommand{\ValPetsMax}{59}

\newcommand{\ValPetsMin}{21}
\newcommand{\ValPetsStepMax}{4.8}
\newcommand{\ValPetsStepMin}{1.7}

\newcommand{\WRhoSizeDir}{-0.86}
\newcommand{\WRhoSizePets}{-0.51}

\newcommand{\WShiftDir}{14.5}
\newcommand{\WShiftDis}{3.6}
\newcommand{\WShiftPets}{7.7}

\newcommand{\WSmallDirP}{34.3}
\newcommand{\WSmallDirU}{47.1}

\newcommand{\WSmallPetsP}{43.4}
\newcommand{\WSmallPetsU}{47.8}
\newcommand{\WorstDirDelta}{+1.64}
\newcommand{\WorstDirP}{p=0.047}
\newcommand{\WorstDirWins}{12/15}

\theoremstyle{thmstyleone}
\newtheorem{proposition}{Proposition}
\theoremstyle{thmstyletwo}
\newtheorem{example}{Example}

\newcommand{\upgrad}{\mathrm{UPGrad}}
\newcommand{\dualproj}{\mathrm{DualProj}}
\newcommand{\gbar}{\bar g}

\newcommand{\suppref}[1]{}

\usepackage{etoolbox}
\makeatletter
\patchcmd{\@maketitle}{\removelastskip\vskip24pt}{\removelastskip\vskip12pt}{}{}
\patchcmd{\@maketitle}{\removelastskip\vskip24pt}{\removelastskip\vskip12pt}{}{}
\makeatother

\begin{document}

\title[Four quantities that come apart in federated dual-cone aggregation]{Conflict,
Feasibility, Faithfulness, Displacement: Four Quantities That Come Apart in Federated
Dual-Cone Aggregation}

\author*[1]{\fnm{Grzegorz} \sur{Gwardys}}\email{grzegorz.gwardys@pw.edu.pl}
\author[2]{\fnm{Karol} \sur{Wojciechowski}}\email{karol.wojciechowski.it@gmail.com}
\author[3]{\fnm{Mateusz} \sur{Gwardys}}\email{m.gwardys@uksw.edu.pl}
\author[2]{\fnm{Bartosz} \sur{{\'S}widerski}}\email{bartosz\_swiderski@sggw.edu.pl}
\affil*[1]{\orgdiv{Faculty of Electronics and Information Technology},
  \orgname{Warsaw University of Technology}, \orgaddress{\city{Warsaw}, \country{Poland}}}
\affil[2]{\orgdiv{Department of Artificial Intelligence, Institute of Information Technology},
  \orgname{Warsaw University of Life Sciences}, \orgaddress{\city{Warsaw}, \country{Poland}}}
\affil[3]{\orgdiv{Institute of Computer Science},
  \orgname{Cardinal Stefan Wyszy{\'n}ski University},
  \orgaddress{\city{Warsaw}, \country{Poland}}}

\abstract{FedFV, FedGH and FedMGDA+ aggregate client updates with multi-objective gradient
aggregators to remove conflict between them. We
examine this rationale with UPGrad, which increases no client's loss to first order, in a
comparison that changes only the direction of FedAvg's update. Four quantities the rationale
treats as one---conflict between client updates, feasibility of their weighted mean, faithfulness
of the transmitted updates to the clients' gradients, and displacement of the update---come apart.
Under Dirichlet partitioning UPGrad improves CIFAR-10 accuracy by $\AccDirDelta$~pp
($\AccDirWins$ seeds, $p<0.001$), only with full participation; on Oxford-IIIT Pet its $\AccPetsDelta$~pp gain vanishes
($\DLTenAccDelta$~pp) once the data our protocol left unused are kept. Varying only the
Dirichlet concentration, the gain grows with heterogeneity, yet the split with the most conflict
shows none. Uniform client weights, blind to conflict, match the CIFAR-10 gain and
exceed the Pets gain; a matched random rotation does neither. No fairness effect survives multiplicity correction,
and on CIFAR-10 the guarantee does not reach the clients' gradients. We state what evaluations
should report.}


\maketitle

\section{Introduction}\label{sec:intro}

Federated learning (FL) trains a shared model across clients that never exchange raw data. In
FedAvg~\citep{mcmahan2017fedavg} each client runs a few epochs of local SGD and the server
averages the resulting models, weighted by local dataset size. When client distributions
differ, local updates drift towards locally good but globally inconsistent
solutions~\citep{karimireddy2020scaffold}. A common diagnosis is that client updates
\emph{conflict}---point in partly opposing directions---so that averaging cancels useful
signal~\citep{wang2021fedfv,zhang2024tackling}.

The same problem arises whenever several losses share one set of parameters, and
multi-objective optimization, multi-task learning (MTL) above all, has produced \emph{gradient
aggregators} for it: rules that combine the per-objective gradients, stacked as the rows of a
Jacobian, into one update while limiting the conflict between them (Sect.~\ref{sec:related}).
With each client's update as one row, the FL server faces exactly this problem: FedFV and FedGH
transfer PCGrad-style projections~\citep{yu2020pcgrad} to federated aggregation, and
FedMGDA+~\citep{hu2022fedmgda} transfers MGDA~\citep{sener2018mgda}, motivated by fairness and
robustness, to obtain a step that no participating client opposes. An aggregator is
\emph{non-conflicting} if its update has a non-negative inner product with every row, so that, to
first order, no objective's loss increases; UPGrad~\citep{quinton2024jacobian} is non-conflicting
and, unlike MGDA, also responds linearly to a rescaling of each row. This makes it the natural
operator for examining the conflict rationale in FL: if conflict between clients is what averaging
gets wrong, an aggregator that provably removes first-order conflict should help.

The rationale runs from conflict to benefit---client updates conflict, so averaging harms some
clients, so an aggregator that removes the conflict should help---and each step of it rests on a
different quantity: how strongly client updates oppose one another (\emph{conflict}); whether
their weighted mean actually violates the non-conflicting condition, so that there is something
to repair (\emph{feasibility}); whether the updates the server receives are the gradients the
guarantee is about (\emph{faithfulness}); and how far the operator finally moves the update once
its change of step length is set aside (\emph{displacement}). The FL methods above are motivated by
the first and evaluated by accuracy; none of them reports the two quantities in between, and only
FedFV separates the direction an aggregator chooses from the change of step norm it introduces,
which otherwise acts as a hidden change of learning rate.

We measure all four for UPGrad, in a comparison in which UPGrad replaces FedAvg's weighted mean
and nothing else changes: both arms use the same client weighting, and UPGrad's step is rescaled
to the norm of the mean. The quantities come apart within each configuration: whether the mean
is feasible depends on the trajectory more than on the data; UPGrad moves the update even where the
mean is feasible; the guarantee holds on the transmitted updates by construction but not on the
clients' gradients after a local epoch; and most of what UPGrad does is a change of length that the
comparison removes.

The accuracy gains UPGrad does produce are small, and the level of conflict does not predict
them. Across
configurations, the one with by far the most conflict shows no detectable gain; within one,
varying only the Dirichlet concentration, the gain grows with heterogeneity, and conflict,
infeasibility and displacement grow with it. Neither gain survives every check: the one on Pets,
where the mean never violates the non-conflicting condition, disappears under a change of data
protocol unrelated to aggregation, and the one on CIFAR-10 when half the clients take part in each
round. Two controls show where the gains can come from: weighting the clients equally, which
ignores conflict, matches UPGrad's gain on CIFAR-10 and exceeds it on Pets, while a random rotation
of the same angle does neither. Our contributions are:
\begin{enumerate}
\item a diagnostic protocol that measures the four quantities separately, distinguishes three
  senses of ``non-conflicting''---a non-negative margin against every transmitted update or every
  client's true gradient, and no client's loss rising after a finite step---and traces the
  operator's displacement to the Gramian by exact counterfactuals (Sects.~\ref{sec:method}
  and~\ref{sec:why});
\item a direction-only comparison of UPGrad and the weighted mean in three configurations, with a
  controlled sweep of Dirichlet heterogeneity, robustness checks, paired statistics, equivalence
  tests and a multiplicity correction (Sect.~\ref{sec:results});
\item two controls, applied to both gains, that separate what UPGrad's direction contributes from
  the reweighting it implies and from the size of the change it makes
  (Sect.~\ref{sec:gains}).

\end{enumerate}
The comparison concerns UPGrad's \emph{direction} at matched step length and weighting, not the
performance of a tuned method (Sect.~\ref{sec:confounds}).

\section{Background and related work}\label{sec:related}

\paragraph{Federated optimization.} FedProx~\citep{li2020fedprox} adds a proximal term to the
local objective and SCAFFOLD~\citep{karimireddy2020scaffold} corrects local steps with control
variates; both act during local training. FedNova~\citep{wang2020fednova} normalizes each
client's update by its number of local steps, since clients that take more steps contribute
longer updates and the plain average then optimizes a different objective. Adaptive federated
optimization~\citep{reddi2021adaptive} treats the averaged client update as a
\emph{pseudo-gradient} for a server-side optimizer. The methods we study act only at the server:
they replace the average of pseudo-gradients by another combination of them.

\paragraph{Gradient aggregation for multiple objectives.} \citet{sener2018mgda} cast MTL as
multi-objective optimization and solve it with MGDA, whose min-norm update is non-conflicting.
PCGrad~\citep{yu2020pcgrad} projects each task gradient onto the normal plane of the gradients it
conflicts with; CAGrad~\citep{liu2021cagrad} maximizes the worst task improvement within a ball
around the average gradient. Neither is non-conflicting in general: PCGrad can increase a task's
loss when there are more than two tasks~\citep{quinton2024jacobian}. \citet{quinton2024jacobian} formalize descent on vector-valued
objectives as \emph{Jacobian descent}, introduce UPGrad, which projects each row of the Jacobian
onto the dual cone of all rows before averaging, and characterize aggregators by whether they are
non-conflicting, linear under scaling and weighted. They also observe that the weights these
aggregators give the rows depend on the Jacobian only through its Gramian, which can be
accumulated block by block; our implementation rests on both facts.

\paragraph{Conflict in federated learning.} FedFV~\citep{wang2021fedfv} links conflict to
fairness: it projects conflicting client updates onto each other's normal planes, in an order
set by the clients' losses and optionally also against earlier rounds' updates, and rescales the
result to the norm of the plain average of the round's updates. FedGH~\citep{zhang2024tackling}
applies PCGrad's pairwise projection to the client updates at the server, without norm
compensation, and reports gains as a plug-in to four FL methods, though not at every client
count of its ablation (on CIFAR-10 it falls below FedAvg at $K=5$ and $K=30$).
FedMGDA+~\citep{hu2022fedmgda} casts FL as multi-objective optimization and combines the
normalized client updates by MGDA's minimum-norm weights, kept within a distance $\epsilon$ of
FedAvg's; the result is a common descent direction for the participating clients. CRAFT~\citep{wang2026craft} takes the update closest to the previous round's
direction subject to equal alignment with every normalized client update, layer by layer, and
argues that this promotes a common-descent structure. We reimplement none of them: our question is what one non-conflicting
operator does to FedAvg's update, not which method performs best. To our knowledge UPGrad has
not been used for client aggregation before.

\section{Method}\label{sec:method}

\subsection{Setting}\label{sec:setting}

$K$ clients hold datasets of sizes $n_k$, $n=\sum_k n_k$, and the global objective is
$F(w)=\sum_k (n_k/n)\,L_k(w)$. At round $t$ the server broadcasts $w_t$; client $k$ trains for
$E$ local epochs of minibatch SGD with momentum at rate $\eta_\ell$, reaching $w_k^{(t)}$, and
returns its pseudo-gradient
\begin{equation}
  g_k \;=\; \bigl(w_t - w_k^{(t)}\bigr)/\eta_\ell .
  \label{eq:pseudo}
\end{equation}
For a single full-batch step, $g_k$ is exactly the gradient of the regularized objective
$\widetilde L_k = L_k + \tfrac{\lambda}{2}\lVert w\rVert^2$ that SGD with weight decay $\lambda$
descends; after an epoch it is a momentum-weighted sum of minibatch gradients along the local
trajectory. We call $g_k$ the \emph{transmitted update} and write $h_k=\nabla L_k(w_t)$ for the
client's true full-batch gradient. The server stacks the $g_k$ as the rows of
$J_t\in\mathbb{R}^{K\times N}$ and steps
\begin{equation}
  w_{t+1} \;=\; w_t - \eta_s\, A(J_t)
  \label{eq:server}
\end{equation}
with an aggregator $A$. With $A$ the sample-weighted mean $\gbar=\sum_k p_k g_k$, $p_k=n_k/n$, and
$\eta_s=\eta_\ell$, Eq.~\eqref{eq:server} is algebraically FedAvg.

\subsection{Dual cone, margins and the two operators}\label{sec:operators}

Let $C=\{d:\ g_k^\top d\ge 0\ \forall k\}$ be the dual cone of the rows and $P_C$ the Euclidean
projection onto it. For an update $d$ we call $g_k^\top d$ client $k$'s (first-order)
\emph{margin}: were $g_k$ the gradient of client $k$'s loss, a step $-\eta d$ would change that loss
by $-\eta\, g_k^\top d$ to first order. An update is \emph{feasible} if it lies in $C$, and a client
with a negative margin is \emph{harmed}. Two non-conflicting operators differ in the order of
projection and averaging:
\begin{equation}
  d_{\upgrad} = \sum_k p_k\, P_C(g_k), \qquad d_{\dualproj} = P_C\Bigl(\sum_k p_k g_k\Bigr).
  \label{eq:operators}
\end{equation}
A feasible mean makes DualProj the identity but not UPGrad, whose $K$ row projections can each
move (Example~\ref{ex:dualproj}, Appendix~\ref{app:proofs}). By the optimality conditions of the
projection, $P_C(g_k)=J^\top(e_k+\mu_k)$ with multipliers $\mu_k\ge0$, so
$d_{\upgrad}=u^\top J$ with effective weights $u=p+\sum_k p_k\mu_k\ge p$
\citep[cf.][Prop.~1]{quinton2024jacobian}: a client gains weight whenever its constraint is
active in the projection of another client's update. For every aggregator we use the weights
depend on $J$ only through the Gramian $G=JJ^\top$, and the server accumulates
$G=\sum_\tau J_\tau J_\tau^\top$ tensor by tensor, as \citet[Eq.~12, Alg.~3]{quinton2024jacobian}
do layer by layer, so it never materializes $J$. TorchJD computes $P_C$ as a slightly
regularized quadratic program, so the feasibility of the computed update is measured, not
guaranteed (Sect.~\ref{sec:binds}).

\subsection{A direction-only comparison}\label{sec:confounds}

UPGrad's output is longer than the weighted mean, by a factor of $\NormRatioMin$--$\NormRatioMax$
in our runs (Table~\ref{tab:displacement}), which at equal $\eta_s$ is a hidden learning-rate
multiplier; we rescale it to the norm of the mean. FedAvg weights clients by $n_k/n$, whereas
TorchJD's aggregators weight uniformly by default; we pass $p_k=n_k/n$ as UPGrad's preference
vector and compute the reference mean with the same weights. The comparison therefore answers one question, (A): at the same
step length and client weighting, does UPGrad's \emph{direction} help? Question (B)---does a
tuned UPGrad-based method outperform tuned alternatives such as FedProx, SCAFFOLD, FedFV or FedGH
at equal computation and communication?---needs a learning-rate search for every method and is
not addressed here. The no-projection arm is the
weighted mean computed in the same gradient path as UPGrad, so the two arms share every numerical
detail but the projection.

\subsection{Measuring the four quantities}\label{sec:measure}

Every round we log, for the weighted mean and for the UPGrad update, all client margins as cosines
$g_k^\top d/(\lVert g_k\rVert\lVert d\rVert)$, the pairwise cosines between client updates, the
cosine between the two updates, the norm ratio and the Gramian; the no-projection arm logs the
margins of its own update, so feasibility can be compared along both trajectories. Three probes
run on selected rounds and evaluate the model without updating it. The \emph{gradient probe} has
every client also return $h_k$, computed over its full local training split, at rounds
$\{1,5,10,20,40,60\}$ (Pets: up to 40), and has it evaluate its own training loss after the step
taken and after the step not taken. On Pets, whose network has BatchNorm, $h_k$ is computed twice:
in \texttt{eval} mode, with the running statistics, and in the mode local training uses, with the
trainable layers normalizing by the statistics of minibatches of the training batch size. The
\emph{loss probe} repeats the loss evaluation at
$\eta\in\{1,\tfrac14,\tfrac1{16},\tfrac1{64}\}\,\eta_s$ (3 seeds, 4 rounds). The
\emph{three-input control} also runs UPGrad on the true gradients $h_k$ and on
$s_k=(\lVert g_k\rVert/\lVert h_k\rVert)\,h_k$, the gradients' directions at the updates' norms.

\subsection{Controls for the mechanism}\label{sec:controls}

Two controls replace UPGrad's direction while keeping one of its properties. Both run in the same
path as UPGrad, with the same reference mean and the same rescaling to its norm, so each differs
from the no-projection arm in one respect. The \emph{uniform-weight control} steps along the
unweighted mean of the updates: the extreme form of the reweighting UPGrad applies (its effective
weights move mass towards small clients, Sect.~\ref{sec:pets}), with nothing that depends on
conflict. The \emph{matched random rotation} turns $\gbar$ by exactly the angle UPGrad would turn
it by on the same Gramian, towards a direction of the span of the updates that is uniformly
random among those orthogonal to $\gbar$ and drawn afresh each round from the run's seed: it keeps
UPGrad's displacement and discards its orientation. For \emph{partial participation}, $m$ of the
$K$ clients are drawn each round from the run's seed and the round index, so that both arms train
on the same clients in every round; the others skip local training, and every client is still
evaluated.

\section{Experimental setup}\label{sec:setup}

\paragraph{Regimes.} (i) \emph{Dirichlet CIFAR-10}~\citep{krizhevsky2009learning}: $K=\KDir$
clients, label proportions drawn from a Dirichlet distribution with
$\alpha=0.5$~\citep{hsu2019dirichlet}; the largest client holds $\SizeSpreadMin$--$\SizeSpreadMax$
times the data of the smallest. (ii) \emph{Disjoint-class CIFAR-10}: $K=\KDis$ clients with two
classes each and no class shared, a purpose-built maximum-conflict split. (iii) \emph{Dirichlet
Oxford-IIIT Pet}~\citep{parkhi2012cats}: $K=\KPets$, $\alpha=0.5$, 37 classes, fine-tuning an
ImageNet-pretrained~\citep{deng2009imagenet} ResNet-18~\citep{he2016resnet} with \texttt{layer4}
and the classifier trainable (8.4\,M of 11.2\,M parameters). On CIFAR-10 we train a LeNet-5-style
CNN~\citep{lecun1998mnist} from scratch (two convolutional and three fully connected layers,
62\,k parameters, no BatchNorm). The three regimes differ in more than heterogeneity, so we also
vary heterogeneity alone: (iv) \emph{controlled heterogeneity}, regime (i) with
$\alpha\in\{0.1,0.5,1,10\}$ and the protocol otherwise unchanged.

\paragraph{Protocol.} Local SGD with momentum $0.9$ and weight decay $5\times10^{-4}$, a fresh
optimizer each round, $\eta_\ell=\eta_s=0.01$ and one local epoch ($E=1$), with batch size
$\BatchCifar$ on CIFAR-10 and $\BatchPets$ on Pets and the last partial batch dropped. An epoch is
on average $\StepsDir$ local steps on Dirichlet CIFAR-10 (range $\StepsDirMin$--$\StepsDirMax$),
$\StepsDis$ on the disjoint split and $\StepsPets$ on Pets (range $\StepsPetsMin$--$\StepsPetsMax$).
All clients participate in every round (except in Sect.~\ref{sec:robust}), for $\RoundsDir$
rounds on CIFAR-10 and $\RoundsPets$ on Pets, in the Flower simulation
engine~\citep{beutel2020flower}. Each client holds out $20\%$ of its data for validation. A seed
sets both the model initialization and the partition, both arms run on the same seeds, and the
main comparisons use $\SeedsDir$ seeds on CIFAR-10 and $\SeedsPets$ on Pets. Each client re-seeds
its generators at the start of every round from the run's seed and its partition index, so it
visits its training split in the same order, with the same augmentation, in every round; with the
last partial batch dropped, the same samples are left out every time. On Pets that is
$\DropSharePets\%$ of the training data, never used for training ($\DropShareDir\%$ on Dirichlet
CIFAR-10), which is why we repeat the Pets comparison with every sample kept
(Sect.~\ref{sec:robust}).

\paragraph{Metrics and statistics.} \emph{Centralized accuracy} is measured on the dataset's test
split; client-level metrics summarize accuracy on the clients' validation splits: unweighted mean
(\emph{client-mean}), mean weighted by split size (\emph{sample-weighted}), minimum
(\emph{worst-client}), mean of the worst $\max(1,\lfloor K/10\rfloor)$ clients
(\emph{worst-decile}), max--min \emph{gap} and standard deviation (\emph{spread}); the last four
are the fairness metrics. All are read at the final round. Comparisons are paired by seed: mean
paired difference, number of seeds favouring UPGrad, two-sided paired $t$-test and exact sign test.
Non-significance is not read as equivalence: we run two one-sided tests (TOST) against
$\pm0.5$~pp, a difference in accuracy we would regard as the smallest worth acting on, with the
matching $90\%$ interval; for the gap and the spread the same number has no such meaning. Other
intervals are $95\%$. The family is seven metrics in three regimes, $\FamilySize$ tests, with
Bonferroni threshold $0.05/\FamilySize=\BonfThreshold$; the seeds are the independent units, and
rounds and clients within a seed are not. Pooled counts are given in round-seeds (one round of one
seed) or client-rounds. On Pets a client's validation split holds $\ValPetsMin$--$\ValPetsMax$
images, so a single prediction moves its accuracy by $\ValPetsStepMin$--$\ValPetsStepMax$~pp:
worst-client and gap differences there are differences of a few predictions.
Configurations and metrics were fixed before the main runs; the seed counts were not. They were
raised from five (three on Pets) after preliminary runs, each final sweep was then run in one
batch, and the seeds the extension added, analysed alone, give the same result\suppref{Sect.~S2}.
We claim no calibrated error rate for the study as a whole: neither the hypotheses nor the analysis
were registered in advance.

\paragraph{Measurement validity.} Effects below one percentage point are smaller than several
artifacts of an FL test bench\suppref{Sect.~S1}. Seeding alone does not make GPU training
reproducible: the simulation returns client replies in completion order and floating-point addition
is not associative, so we aggregate in a fixed order (by partition id), which makes repeated runs
identical. FedAvg's weight average and our gradient-path mean are algebraically identical but
numerically different; no offset between them is detected in any regime, and both arms of every
comparison run in the same path. Flower builds the simulation's software environment anew for each
run, and before we pinned it, package versions changed between runs: every run we report records
its environment, the main sweeps share one, and all later runs (Sects.~\ref{sec:robust},
\ref{sec:alpha} and~\ref{sec:ctrlres}) used it pinned\suppref{Sect.~S1.5}.

\section{Accuracy and fairness}\label{sec:results}

\subsection{Main comparison}\label{sec:main}

\begin{table}[!htb]
\caption{Centralized test accuracy, weighted mean vs.\ UPGrad, paired by seed. $\Delta$ is
UPGrad minus mean; \emph{wins} counts seeds on which UPGrad scored higher; $t$ and \emph{sign} are
two-sided paired $t$- and exact sign tests; the $90\%$ CI and $p_{\rm TOST}$ belong to the
equivalence test against $\pm0.5$~pp.}\label{tab:accuracy}
  \setlength{\tabcolsep}{4pt}
  \footnotesize
  \begin{tabular*}{\textwidth}{@{\extracolsep{\fill}}lccccccccc}
    \toprule
    & & \multicolumn{2}{c}{accuracy (\%)} & & & \multicolumn{2}{c}{$p$} & & \\
    \cmidrule(lr){3-4}\cmidrule(lr){7-8}
    Regime & $n$ & mean & UPGrad & $\Delta$ (pp) & wins & $t$ & sign & $90\%$ CI & $p_{\rm TOST}$ \\
    \midrule
    Disjoint   & 15 & 38.94 & 38.71 & $-0.23$ & 6/15 & 0.362 & 0.607 & $[-0.66, +0.20]$ & 0.141 \\
    Dirichlet  & 15 & 56.72 & 57.51 & $+0.78$ & 14/15 & $<$0.001 & $<$0.001 & $[+0.55, +1.01]$ & 0.974 \\
    Pets       & 10 & 80.66 & 81.18 & $+0.52$ & 10/10 & $<$0.001 & 0.002 & $[+0.33, +0.71]$ & 0.587 \\
    \bottomrule
  \end{tabular*}

\end{table}

\begin{figure}[!htb]
\centering
\includegraphics[width=0.62\textwidth]{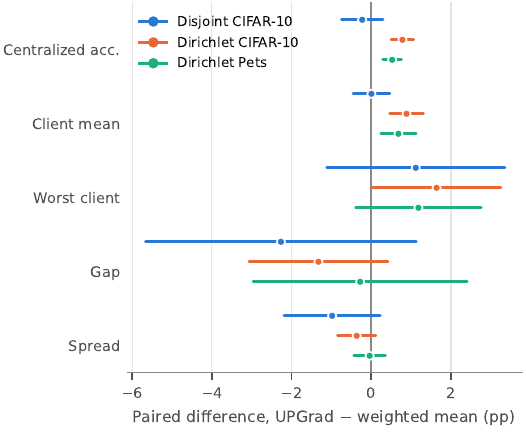}
\caption{Paired differences (UPGrad minus mean) at the final round, with $95\%$ confidence
intervals. Positive favours UPGrad except for the gap and the spread, where negative does.}
\label{fig:effects}
\end{figure}

Under Dirichlet partitioning UPGrad improves centralized accuracy by $\AccDirDelta$~pp on CIFAR-10
($\AccDirWins$ seeds, $\AccDirP$) and $\AccPetsDelta$~pp on Pets ($\AccPetsWins$, $\AccPetsP$); on
the disjoint split no difference is detected ($\AccDisDelta$~pp, $\AccDisWins$, $\AccDisP$)
(Table~\ref{tab:accuracy}, Fig.~\ref{fig:effects}). The exact sign test gives the same picture
($\AccDirSignP$, $\AccPetsSignP$, $\AccDisSignP$), so the result does not rest on normality. The
Pets gain, however, does not survive keeping the training samples the protocol leaves out
(Sect.~\ref{sec:robust}).
For centralized accuracy equivalence is established in no regime, for three different reasons: on Dirichlet CIFAR-10 the $90\%$
interval $\AccDirCI$ lies above the margin, so the effect exceeds it; on Pets ($\AccPetsCI$) a
positive effect is shown but its relation to the margin is not; on the disjoint split
($\AccDisCI$) the test is simply inconclusive. Among the other metrics, client-mean accuracy on the
disjoint split is equivalent within the margin ($\EqDisMeanCI$, $\EqDisMeanP$).

The fairness metrics lean the same way but are not resolved. All $\FairCells$ fairness cells
(four metrics in three regimes) favour UPGrad in their point estimates, and worst-client accuracy
moves $\WorstDirDelta$~pp under Dirichlet CIFAR-10 ($\WorstDirWins$, $\WorstDirP$). Of the
$\FamilySize$ tests, $\FamilySig$ reach $p<0.05$, two of them fairness cells on Dirichlet CIFAR-10
($\WorstDirP$ and $\DecileDirP$); the $\FamilyBonf$ that survive the correction are all accuracy
metrics (Table~\ref{tab:family}, Appendix~\ref{app:tables}).

\paragraph{An ordering across non-comparable configurations.} The disjoint split produces by
far the most conflict (Sect.~\ref{sec:conflict}) and is the one main regime without a detected
gain. The configurations also differ in client count, local step count, architecture and
initialization, so this ordering estimates no effect of heterogeneity; it challenges only the
expectation that stronger conflict necessarily yields a larger benefit from conflict-averse
aggregation, and Sect.~\ref{sec:alpha} varies heterogeneity alone.

\subsection{Robustness}\label{sec:robust}

\begin{table}[!htb]
\caption{Pets, UPGrad minus mean at round~40, under both \texttt{drop\_last} settings on the same
ten seeds; \emph{change} is the paired difference of the two differences.}\label{tab:droplast10}
  \setlength{\tabcolsep}{4pt}
  \footnotesize
  \begin{tabular*}{\textwidth}{@{\extracolsep{\fill}}lcccccccc}
    \toprule
    & \multicolumn{3}{c}{\texttt{drop\_last=True}} & \multicolumn{3}{c}{\texttt{drop\_last=False}} & \multicolumn{2}{c}{change} \\
    \cmidrule(lr){2-4}\cmidrule(lr){5-7}\cmidrule(lr){8-9}
    Metric & $\Delta$ & wins & $p$ & $\Delta$ & wins & $p$ & $\Delta$ & $p$ \\
    \midrule
    centralized accuracy & $+0.52$ & 10/10 & $<$0.001 & $+0.07$ & 6/10 & 0.320 & $-0.46$ & $<$0.001 \\
    client-mean accuracy & $+0.68$ & 9/10 & 0.007 & $+0.35$ & 8/10 & 0.006 & $-0.33$ & 0.219 \\
    sample-weighted accuracy & $+0.63$ & 9/10 & 0.004 & $+0.31$ & 7/10 & 0.016 & $-0.32$ & 0.177 \\
    worst-client accuracy & $+1.18$ & 5/10 & 0.124 & $-0.28$ & 2/10 & 0.677 & $-1.46$ & 0.136 \\
    worst-decile accuracy & $+0.73$ & 6/10 & 0.248 & $-0.22$ & 4/10 & 0.645 & $-0.95$ & 0.034 \\
    max--min gap & $-0.28$ & 5/10 & 0.820 & $+0.53$ & 2/10 & 0.478 & $+0.81$ & 0.532 \\
    spread across clients & $-0.04$ & 6/10 & 0.821 & $+0.05$ & 4/10 & 0.744 & $+0.09$ & 0.666 \\
    \bottomrule
  \end{tabular*}

\end{table}

\paragraph{Every training sample kept.} With the last partial batch dropped and the data order
fixed, the Pets protocol never trains on $\DropSharePets\%$ of the training data
(Sect.~\ref{sec:setup}). Repeated on the same $\SeedsPets$ seeds with every sample kept, which
raises a client's local steps to $\DLTenSteps$ per epoch and FedAvg's final accuracy by
$\DLTenGainMean$~pp, the comparison loses its gain (Table~\ref{tab:droplast10}): centralized
accuracy differs by $\DLTenAccDelta$~pp ($\DLTenAccWins$ seeds, $\DLTenAccP$), equivalent to zero
within $\pm0.5$~pp (TOST $\DLTenAccTOSTP$), and the change from the original protocol,
$\DLTenAccChange$~pp ($\DLTenAccChangeP$), is itself detected. Client-mean and sample-weighted
accuracy keep about half their gain, neither below the Bonferroni threshold of the main family,
and the fairness metrics show nothing. A lower test loss remains: UPGrad's is lower in every seed
at every round ($\TrajPetsAllLossLast\times10^{-3}$ at round~40, against
$\TrajPetsLossLast\times10^{-3}$ with the tail dropped), while its accuracy lead of
$\TrajPetsAllAccTen$~pp at round~10 closes by round~40 (Fig.~\ref{fig:trajectory}). The Pets
accuracy gain is therefore a property of the protocol as much as of the aggregator; why discarding
the same samples in every round makes the final accuracy depend on the aggregator, we have not
established.

\paragraph{Partial participation.} With $\PPClients$ of the $20$ clients drawn in each round
(Dirichlet CIFAR-10, seeds 1--10; both arms train the same clients in all $\PPRoundSeeds$
round-seeds), the gain is not detected: centralized accuracy $\PPAcc$~pp ($\PPAccWins$ seeds,
$\PPAccP$), equivalent to zero within $\pm0.5$~pp (TOST $\PPAccTOSTP$), against $\PPFullAcc$~pp on
the same seeds with every client in every round; neither the test loss nor any fairness metric
differs detectably. UPGrad also changes the update about half as much (rotation $\PPRot^\circ$
against $\PPFullRot^\circ$, weight mass moved $\PPShift\%$ against $\PPFullShift\%$), and the mean
is infeasible along its trajectory in $\PPInf\%$ of round-seeds against
$\PPFullInf\%$\suppref{Sect.~S5}. The Dirichlet CIFAR-10 gain therefore depends on full
participation in our setting; why UPGrad reweights less when half the clients take part, we have
not established.

\subsection{Controlled heterogeneity}\label{sec:alpha}

\begin{table}[!htb]
\caption{Dirichlet CIFAR-10 at four concentrations ($K=20$, seeds 1--10; at $\alpha=0.5$ the
first ten seeds of the main sweep), UPGrad minus mean at round~60; lower $\alpha$ is more
heterogeneous. \emph{\% of floor}: $\bar c$ as a percentage of $-1/(K-1)$, negative when the
clients agree on average. \emph{Infeasible}: share of round-seeds in which the weighted mean
of the transmitted updates would harm some client to first order, along each arm's trajectory.
\emph{Align}: median $\cos(g_k,h_k)$ over clients (UPGrad arm); $\cos(d(g),d(h))$: UPGrad's
direction on the transmitted updates against its direction on the true gradients.
\emph{Removed by equal norms}: share of the rotation that setting all update norms equal removes
(the Gramian counterfactual of Sect.~\ref{sec:trigger});
\emph{weight mass moved}: median total-variation distance between UPGrad's implied client weights
and the sample-size weights.}\label{tab:alpha}
  \setlength{\tabcolsep}{4pt}
  \footnotesize
  \begin{tabular*}{\textwidth}{@{\extracolsep{\fill}}lcccc}
    \toprule
    Dirichlet $\alpha$ & 0.1 & 0.5 & 1.0 & 10.0 \\
    \midrule
    $\Delta$ accuracy (pp, wins) & $+0.85$ (7/10) & $+0.92$ (10/10) & $+0.25$ (9/10) & $+0.06$ (5/10) \\
    \quad $p$ ($t$ / sign) & 0.062 / 0.344 & $<$0.001 / 0.002 & 0.016 / 0.021 & 0.332 / 1.000 \\
    $\Delta$ client-mean (pp, wins) & $+2.60$ (10/10) & $+0.93$ (10/10) & $+0.20$ (7/10) & $+0.12$ (5/10) \\
    \quad $p$ ($t$) & $<$0.001 & 0.009 & 0.097 & 0.182 \\
    $\Delta$ worst client (pp) & $+2.91$ & $+1.39$ & $+0.01$ & $+1.00$ \\
    $\Delta$ gap (pp) & $+0.64$ & $-0.89$ & $+0.16$ & $-0.97$ \\
    \midrule
    conflict: $\bar c$ & $-0.014$ & $+0.005$ & $+0.029$ & $+0.081$ \\
    \quad \% of floor & $27$ & $-10$ & $-54$ & $-154$ \\
    \quad conflicting pairs (\%) & 60 & 57 & 55 & 50 \\
    feasibility: infeasible (\%), UPGrad run & 98 & 84 & 36 & 0.3 \\
    \quad no-projection run & 65 & 7.7 & 0.2 & 0.3 \\
    faithfulness: align, round 1 & 0.85 & 0.85 & 0.87 & 0.95 \\
    \quad align, round 60 & 0.53 & 0.29 & 0.24 & 0.26 \\
    \quad $\cos(d(g),d(h))$ & 0.28 & 0.22 & 0.23 & 0.30 \\
    displacement: rotation ($^\circ$) & 28.2 & 16.2 & 8.8 & 2.4 \\
    \quad norm ratio & 1.76 & 1.68 & 1.57 & 1.41 \\
    \quad removed by equal norms (\%) & 7 & 17 & 24 & 0 \\
    \quad weight mass moved (\%) & 29.1 & 14.1 & 6.6 & 1.7 \\
    \midrule
    local steps: mean (range) & 62 (1--240) & 62 (7--165) & 62 (26--122) & 62 (47--78) \\
    largest / smallest client & 39.6 & 4.8 & 3.1 & 1.5 \\
    \bottomrule
  \end{tabular*}

\end{table}

Only the Dirichlet concentration changes here: CIFAR-10 with $K=20$, the same model, protocol and
$60$ rounds, at $\alpha\in\{0.1,1,10\}$ on seeds 1--10, with $\alpha=0.5$ taken from the main sweep
(Table~\ref{tab:alpha}). The partition changes more than label skew: as $\alpha$ falls, client
sizes become more unequal (largest to smallest client $\AlphaTenSpread$ at $\alpha=10$,
$\AlphaTenthSpread$ at $\alpha=0.1$, median over seeds) and so do local step counts
($\AlphaTenStepsMin$--$\AlphaTenStepsMax$ against $\AlphaTenthStepsMin$--$\AlphaTenthStepsMax$).

\paragraph{Accuracy.} The gain falls as $\alpha$ rises (Table~\ref{tab:alpha}). Spearman's
correlation between $\alpha$ and the per-seed difference is $\AlphaTrendAccRho$ for centralized
accuracy ($\AlphaTrendAccP$) and $\AlphaTrendMeanRho$ for client-mean accuracy ($\AlphaTrendMeanP$;
permutation tests, as the seeds are not paired across $\alpha$\suppref{Sect.~S4}). Taken one
$\alpha$ at a time, only $\alpha=0.5$ passes the Bonferroni threshold $\AlphaBonf$ for the four
accuracy tests. At $\alpha=0.1$ the mean difference is as large as at $0.5$ but varies from seed
to seed, while client-mean accuracy rises on every seed; at $\alpha=10$ the difference is
equivalent to zero within $\pm0.5$~pp (TOST $\AlphaTenAccTOSTP$). Two of the eight fairness cells
have $p<0.05$, both at $\alpha=10$, where the operator changes least, and neither survives a
Bonferroni correction over the eight.

\paragraph{The four quantities.} Along this axis they move together, though not in proportion.
From $\alpha=10$ to $\alpha=0.1$ the share of conflicting client pairs rises only from
$\AlphaTenPairs\%$ to $\AlphaTenthPairs\%$, while the infeasibility of the mean, UPGrad's rotation
and the client-weight mass it moves grow by an order of magnitude or more. Faithfulness does not
follow this order: the median alignment is $\AlphaTenthAlignFirst$--$\AlphaTenAlignFirst$ at
round~1 for every $\alpha$ and by round~60 falls to $\AlphaOneAlignLast$--$\AlphaHalfAlignLast$ for
$\alpha\ge0.5$ but only to $\AlphaTenthAlignLast$ at $\alpha=0.1$. At $\alpha=1$ the mean is
infeasible in $\AlphaOneInfMean\%$ of round-seeds of its own run but in $\AlphaOneInfUp\%$ along
UPGrad's, the trajectory dependence of Sect.~\ref{sec:binds}.

\paragraph{What the sweep shows.} Within one configuration the expectation questioned in
Sect.~\ref{sec:main} holds: more heterogeneity, more benefit. Because conflict, infeasibility,
displacement, client sizes and local step counts all change with $\alpha$, the sweep cannot say
which of them carries the benefit. Across configurations the expectation fails: the disjoint split
has more conflict than any $\alpha$ here ($\FloorAvgDis\%$ of its floor) but no gain, and along
UPGrad's trajectory its mean is infeasible in only $\InfDisPct\%$ of round-seeds, against
$\AlphaTenthInfUp\%$ at $\alpha=0.1$ and $\InfDirPct\%$ under the main Dirichlet split. Over these
CIFAR-10 settings the gain thus follows the infeasibility of the mean more closely than conflict;
that is an ordering of five settings, not a fitted relation, and the original-protocol Pets gain
does not fit it. At $\alpha=0.5$, uniform client weights reproduce both the gain and the
infeasibility along UPGrad's trajectory (Sect.~\ref{sec:ctrlres}), so the two may share a
source.

\section{What the operator does}\label{sec:why}

\subsection{Conflict}\label{sec:conflict}

\begin{figure}[!htb]
\centering
\includegraphics[width=\textwidth]{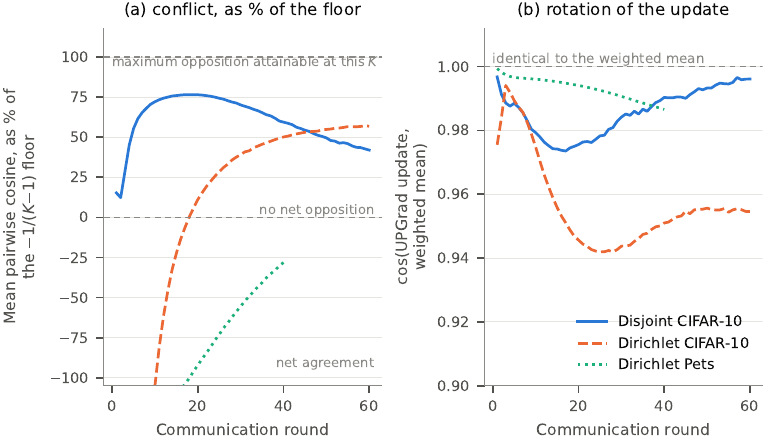}
\caption{Seed-averaged, per round. \textbf{(a)} Mean pairwise cosine as a percentage of the floor
$-1/(K-1)$; negative values mean a net agreeing population. \textbf{(b)} Cosine between the UPGrad
update and the weighted mean: the rotation that remains once the step norm is equalized.}
\label{fig:mechanism}
\end{figure}

For unit vectors $u_k=g_k/\lVert g_k\rVert$ with mean pairwise cosine $\bar c$,
$0\le\lVert\sum_k u_k\rVert^2 = K + K(K-1)\,\bar c$, so
\begin{equation}
  \bar c \;\ge\; -\frac{1}{K-1}.
  \label{eq:floor}
\end{equation}
The floor is $\FloorFive$ for $K=5$ and $\FloorTwenty$ for $K=20$: a negative mean cosine among
many clients is necessarily small, and comparing it with $-1$ makes conflict look absent when it is
not. Against the floor, conflict is substantial (Table~\ref{tab:conflict},
Fig.~\ref{fig:mechanism}a). The disjoint split averages $\bar c=\CosAvgDis$, $\FloorAvgDis\%$ of
the floor, peaks at $\FloorPeakDis\%$, and $\PairsDis\%$ of its client pairs conflict in the
average round. Both Dirichlet regimes are net agreeing on average ($\CosAvgDir$ and $\CosAvgPets$),
yet $\PairsDir\%$ and $\PairsPets\%$ of their pairs conflict in a typical round.

\begin{table}[!htb]
\caption{Mean pairwise cosine between client updates, pooled over every round and seed of the
UPGrad arm ($n$ = rounds $\times$ seeds). \textsc{avg} is the typical round, \textsc{peak} the
most conflicted one, \% floor the peak as a share of $-1/(K-1)$, and \textsc{min pair} the most
opposed pair observed.}\label{tab:conflict}
  \setlength{\tabcolsep}{4pt}
  \footnotesize
  \begin{tabular*}{\textwidth}{@{\extracolsep{\fill}}lcccccc}
    \toprule
    & & & \multicolumn{2}{c}{mean pairwise $\cos$} & & \\
    \cmidrule(lr){4-5}
    Regime & $K$ & $n$ & \textsc{avg} & \textsc{peak} & \% floor & \textsc{min pair} \\
    \midrule
    Disjoint   & 5  & 900 & $-0.151$ & $-0.202$ & \textbf{81\%} & $-0.307$ \\
    Dirichlet  & 20 & 900 & $+0.005$ & $-0.032$ & 61\% & $-0.650$ \\
    Pets       & 20 & 400 & $+0.051$ & $+0.011$ & --- & $-0.348$ \\
    \bottomrule
  \end{tabular*}

\end{table}

\subsection{Feasibility}\label{sec:binds}

Under Dirichlet CIFAR-10 the weighted mean would harm at least one client in $\InfDir$ of $\NDir$
round-seeds ($\InfDirPct\%$), up to $\MaxHarmDir$ of $20$ at once; under disjoint labels in
$\InfDis$ ($\InfDisPct\%$); on Pets in none (Table~\ref{tab:harmed}). UPGrad's update is feasible
in all $\NAll$ round-seeds. These counts describe the states UPGrad training visits; along the
no-projection trajectory the mean is infeasible in only $\TrajDir$ of $\NDir$ round-seeds under
Dirichlet CIFAR-10 ($\TrajDirPct\%$) and $\TrajDis$ under disjoint labels. Most of the binding is
thus induced by the trajectory UPGrad follows, not by averaging on this data.

Feasibility does not order the outcomes across regimes: on Pets the mean is never infeasible, yet
the original protocol shows a gain there, which therefore cannot come from repairing violations. One implication does hold: if every pair of rows has a non-negative inner
product, every row lies in $C$ and UPGrad returns the mean unchanged, so an infeasible mean requires
a conflicting pair. The converse fails, and our regimes sit on exactly that gap.

\begin{table}[!htb]
\caption{Feasibility of the weighted mean and of the UPGrad update, pooled over every round and
seed. \textsc{infeasible} counts round-seeds in which at least one client has a negative margin;
\textsc{worst margin} is the minimum cosine margin observed.}\label{tab:harmed}
  \setlength{\tabcolsep}{4pt}
  \footnotesize
  \begin{tabular*}{\textwidth}{@{\extracolsep{\fill}}lccc}
    \toprule
     & Disjoint & Dirichlet & Pets \\
    \midrule
    clients $K$                & 5 & 20 & 20 \\
    $n$ (rounds $\times$ seeds) & 900 & 900 & 400 \\
    \midrule
    infeasible: mean           & 57 (6\%) & \textbf{750 (83\%)} & 0 \\
    infeasible: projected      & 0 & 0 & 0 \\
    max clients harmed         & 1/5 & 9/20 & 0/20 \\
    \midrule
    worst margin: mean         & $-0.099$ & $-0.216$ & $+0.032$ \\
    worst margin: projected    & $+0.160$ & $+0.004$ & $+0.150$ \\
    \bottomrule
  \end{tabular*}

\end{table}

\subsection{Faithfulness}\label{sec:faithful}

On the transmitted updates UPGrad's guarantee holds by construction; whether it says anything
about the clients' losses depends on how well $g_k$ stands in for $h_k$ (Table~\ref{tab:truegrad}).
On CIFAR-10 the median alignment $\cos(g_k,h_k)$ falls from $\AlignDisFirst$ (disjoint) and
$\AlignDirFirst$ (Dirichlet) at round~1 to $\AlignDisLast$ and $\AlignDirLast$ at round
$\ProbeLastDir$; against the regularized gradient $\nabla\widetilde L_k$, which the updates
descend, the seed-averaged medians differ by at most $\AlignRegDiff$. On the true gradients (\texttt{eval} mode on Pets) the guarantee does not transfer:
UPGrad leaves some client harmed in $\UnrepDis$ of $\ProbedDis$ probed round-seeds under disjoint labels and $\UnrepDir$ of
$\ProbedDir$ under Dirichlet. On Pets, BatchNorm makes ``the local gradient'' ambiguous: the
alignment is $\AlignPetsLast$ against the \texttt{eval}-mode function but $\AlignPetsTrainLast$
against the train-mode function local training descends; there the mean harms nobody on either
reading.

\begin{table}[!htb]
\caption{Transmitted updates vs.\ true gradients at three probe rounds (mean $\pm$ sd over
seeds). \textsc{align} is the median $\cos(g_k,h_k)$ over clients, against the \texttt{eval}-mode
function and, for Pets, the train-mode function local training differentiates (the CIFAR-10 model
has no BatchNorm). The last two columns count harmed clients for the mean $\to$ UPGrad, on the
transmitted updates and on the true gradients.}
\label{tab:truegrad}
  \setlength{\tabcolsep}{4pt}
  \footnotesize
  \begin{tabular*}{\textwidth}{@{\extracolsep{\fill}}lccccc}
    \toprule
    & & \multicolumn{2}{c}{\textsc{align}, against the} & \multicolumn{2}{c}{harmed, mean $\to$ proj.} \\
    \cmidrule(lr){3-4}\cmidrule(lr){5-6}
    Regime & Round & \texttt{eval} function & train-mode function & transmitted & true \\
    \midrule
    \multirow{3}{*}{Disjoint}
      & 1  & $0.66 \pm 0.04$ & same function & $0.0 \to 0.0$ & $2.1 \to 2.5$ \\
      & 20 & $0.63 \pm 0.06$ & same function & $0.2 \to 0.0$ & $2.3 \to 1.5$ \\
      & 60 & $\mathbf{0.38 \pm 0.04}$ & same function & $0.0 \to 0.0$ & $\mathbf{2.1 \to 1.9}$ \\
    \midrule
    \multirow{3}{*}{Dirichlet}
      & 1  & $0.85 \pm 0.03$ & same function & $0.1 \to 0.0$ & $7.3 \to 6.4$ \\
      & 20 & $0.40 \pm 0.03$ & same function & $2.0 \to 0.0$ & $6.7 \to 0.2$ \\
      & 60 & $\mathbf{0.28 \pm 0.03}$ & same function & $4.5 \to 0.0$ & $\mathbf{9.8 \to 3.5}$ \\
    \midrule
    \multirow{3}{*}{Pets}
      & 1  & $0.74 \pm 0.01$ & $0.96 \pm 0.00$ & $0.0 \to 0.0$ & $0.0 \to 0.0$ \\
      & 20 & $0.64 \pm 0.02$ & $0.95 \pm 0.00$ & $0.0 \to 0.0$ & $0.0 \to 0.0$ \\
      & 40 & $\mathbf{0.57 \pm 0.01}$ & $\mathbf{0.95 \pm 0.01}$ & $0.0 \to 0.0$ & $\mathbf{0.0 \to 0.0}$ \\
    \bottomrule
  \end{tabular*}

\end{table}

The norms are not the problem: a positive rescaling $g_k=a_kh_k$ leaves $C$ and the sign of every
margin unchanged, and with a constant gradient momentum alone would make $g_k$ about
$\MomentumFactor$ times $h_k$ over $\StepsDir$ steps. The directions are. Run on the true gradients,
UPGrad returns a direction whose cosine with the one it returns on the transmitted updates
averages $\DdDisMin$--$\DdDisMax$ over the probe rounds on the disjoint split and
$\DdDirMin$--$\DdDirMax$ under Dirichlet CIFAR-10; on Pets, against the \texttt{eval}-mode
gradients, it falls from $\DdPetsMax$ (round~5) to $\DdPetsLast$ (round~40). Run on $s_k$, which
has the gradients' directions and the updates' norms, it reproduces its output on the gradients to
a cosine of at least $\ThirdInputMin$ (Table~\ref{tab:scale}), so the norms do not cause this;
and the plain weighted mean shows the same gap ($\PlainCosDis$ and $\PlainCosDir$ at round~1), so
the operator does not either. The server inherits the gap from what it is sent.

\subsection{Realized loss change}\label{sec:loss}

A non-negative margin is a first-order statement. The measured change of client $k$'s loss after
the step $-\eta d$ splits as
\begin{equation}
  \Delta L_k = -\eta\, g_k^\top d \;-\; \eta\,(h_k-g_k)^\top d \;+\; R(\eta,d),
  \label{eq:split}
\end{equation}
the margin's prediction, the error of using the transmitted update in place of the gradient, and
the Taylor remainder. Both errors vanish as $\eta\to0$, but $R=O(\eta^2)$ disappears from
$\Delta L_k/\eta$ while the substitution error in the predicted directional derivative does not.
After a local epoch the substitution error is dominated by scale---$g_k$ is tens to hundreds of
times longer than $h_k$, so the margin overstates the predicted change by that factor
(Table~\ref{tab:deltal})---and only the margin's sign carries information; that sign is what the
counts below test. The loss probe separates the two errors (Table~\ref{tab:objective}). With a single full-batch local step,
where $g_k$ is the gradient of $\widetilde L_k$, shrinking the step $\StepShrinkFB$-fold removes
every observed loss increase under UPGrad's step ($\LadderFB$ of $\LadderNFB$ client-rounds). After
a local epoch it does not: $\LadderDis$ of $\LadderNDis$ on the disjoint split and $\LadderDir$ of
$\LadderNDir$ under Dirichlet CIFAR-10. At the step actually used UPGrad raises some client's loss
in $\RiseRoundsDisProj$ of $\ProbeRoundsDis$ probe rounds (disjoint) and $\RiseRoundsDirProj$ of
$\ProbeRoundsDir$ (Dirichlet), against all of them for the mean, and raises fewer clients' losses
than the mean would ($\RiseMeanDirProj$ against $\RiseMeanDirMean$ per round under Dirichlet).
Counting on $L_k$ or on $\widetilde L_k$ changes the sign of at most $\SignFlipsMax$ client-rounds.

\begin{table}[!htb]
\caption{Client-rounds in which UPGrad's step raises the client's loss, as the step shrinks from
$\eta_s$ to $\eta_s/64$, counted on the criterion $L_k$ and on the regularized objective
$\widetilde L_k$, with the number of client-rounds whose sign differs.}
\label{tab:objective}
  \setlength{\tabcolsep}{4pt}
  \footnotesize
  \begin{tabular*}{\textwidth}{@{\extracolsep{\fill}}lllcccc}
    \toprule
    & & & \multicolumn{4}{c}{step, as a fraction of $\eta_s$} \\
    \cmidrule(lr){4-7}
    Configuration & $n$ & objective & $1$ & $1/4$ & $1/16$ & $1/64$ \\
    \midrule
    Disjoint, 1 full-batch step  &  60 & $L_k$ & 9 & 1 & 0 & 0 \\
                                 &     & $\widetilde L_k$ & 9 & 1 & 0 & 0 \\
                                 &     & \quad sign flips & 0 & 0 & 0 & 0 \\
    \addlinespace[2pt]
    Disjoint, 1 epoch            &  60 & $L_k$ & 24 & 18 & 18 & 18 \\
                                 &     & $\widetilde L_k$ & 24 & 18 & 18 & 18 \\
                                 &     & \quad sign flips & 0 & 0 & 0 & 0 \\
    \addlinespace[2pt]
    Dirichlet, 1 epoch           & 240 & $L_k$ & 29 & 33 & 34 & 34 \\
                                 &     & $\widetilde L_k$ & 30 & 33 & 34 & 37 \\
                                 &     & \quad sign flips & 1 & 0 & 0 & 3 \\
    \bottomrule
  \end{tabular*}

\end{table}

Exact gradients and an arbitrarily small step are not sufficient either:

\begin{proposition}\label{prop:zero}
There are differentiable objectives and a point at which UPGrad, given the exact gradients,
returns an update against which every margin is non-negative and yet two of the losses increase
for every step length $\eta>0$.
\end{proposition}

At a margin of exactly zero the second-order term decides; the construction (Appendix~\ref{app:proofs})
uses three quadratics. It needs a
degenerate configuration---for UPGrad a margin vanishes only if $P_C(g_k)=0$ and $g_k$ is orthogonal
to every other projected row---so it is a qualification rather than a practical failure: with exact
gradients and strictly positive margins a small enough step decreases every loss, which is the
condition under which ``no objective increases'' for a non-conflicting
aggregator~\citep[Sect.~2.2]{quinton2024jacobian} holds.

\subsection{Displacement}\label{sec:trigger}

Most of what UPGrad does here is a change of length. For a symmetric population---equal norms,
equal weights and a common pairwise cosine $c$ with $-1/(K-1)<c<0$---it is a scalar multiple of
the mean (Proposition~\ref{prop:symmetric}, Appendix~\ref{app:proofs}),
\begin{equation}
  d_{\upgrad} \;=\; \frac{1-c}{1+(K-2)c}\;\gbar ,
  \label{eq:symmetric}
\end{equation}
and applied round by round to the measured mean cosine it predicts the measured norm ratio to a
median relative error of $\SymErrDis\%$ on the near-symmetric disjoint split and $\SymErrDir\%$
under Dirichlet CIFAR-10 (Table~\ref{tab:symmetric}). Our protocol removes this length change, and
what remains is a rotation of median $\RotMedMin$--$\RotMedMax^\circ$ by regime and at most
$\RotWorstDeg^\circ$ (Table~\ref{tab:displacement}, Fig.~\ref{fig:mechanism}b).

\begin{table}[!htb]
\caption{Displacement of the UPGrad update from the weighted mean, pooled over every round and seed.
\emph{Top:} cosine to the mean and norm ratio (removed by our rescaling), mean $\pm$ sd and range.
\emph{Bottom:} mean rotation in degrees when one asymmetry of each round's Gramian is
removed---unequal weights $p$, unequal norms $\mathrm{diag}(G)$, or dispersion of the pairwise
cosines---with the share of the measured rotation the removal eliminates.}
\label{tab:displacement}
  \setlength{\tabcolsep}{4pt}
  \footnotesize
  \begin{tabular*}{\textwidth}{@{\extracolsep{\fill}}lccc}
    \toprule
    Regime & $n$ & $\cos$ vs.\ mean & Norm ratio \\
    \midrule
    Disjoint   & 900 & $0.986 \pm 0.011$ & $2.15 \pm 0.45$ \\
               &     & $[0.940,\,0.999]$ & $[1.07,\,3.02]$ \\
    \addlinespace[2pt]
    Dirichlet  & 900 & $0.957 \pm 0.018$ & $1.67 \pm 0.34$ \\
               &     & $[0.906,\,0.999]$ & $[1.05,\,2.35]$ \\
    \addlinespace[2pt]
    Pets       & 400 & $0.993 \pm 0.004$ & $1.36 \pm 0.17$ \\
               &     & $[0.980,\,1.000]$ & $[1.01,\,1.75]$ \\
    \bottomrule
  \end{tabular*}

\par\vspace{8pt}
  \setlength{\tabcolsep}{4pt}
  \footnotesize
  \begin{tabular*}{\textwidth}{@{\extracolsep{\fill}}lccc}
    \toprule
    Mean rotation ($^\circ$) & Disjoint & Dirichlet & Pets \\
    \midrule
    measured        & 9.05 & 16.47 & 6.44 \\
    \midrule
    $-$ weights     & 9.05 (0\%) & 11.25 (32\%) & 5.83 (9\%) \\
    $-$ norms       & 4.11 (55\%) & 13.62 (17\%) & 6.05 (6\%) \\
    $-$ cosines     & 6.23 (31\%) & 10.69 (35\%) & 0.00 (100\%) \\
    \midrule
    $-$ all three   & 0.00 & 0.00 & 0.00 \\
    \bottomrule
  \end{tabular*}

\end{table}

Because the rotation is an exact function of the Gramian and the weights---with $d=u^\top J$ and
$\gbar=p^\top J$, $\cos(d,\gbar)=u^\top Gp/\sqrt{(u^\top Gu)(p^\top Gp)}$, and $u$ is a function of
$G$ and $p$---its sources can be found by counterfactual rather than by correlation: we remove one
asymmetry from each round's measured Gramian and recompute the angle exactly
(Table~\ref{tab:displacement}, bottom). Which asymmetry matters depends on the regime. On the
disjoint split the client weights contribute nothing, shards being equal, and unequal update norms
account for $\RotShareDisNorms\%$; under Dirichlet CIFAR-10 all three contribute, cosine
dispersion most ($\RotShareDirCos\%$). The shares do not sum to $100\%$ because the asymmetries
interact. No scalar summary we logged determines the rotation: on Pets the mean cosine is positive
in every round and the mean rotation is still $\RotMeanPets^\circ$, because individual pairs
conflict (Example~\ref{ex:meancos}).

\section{Where the gains come from}\label{sec:gains}

Two gains remain to be explained, each under a condition: under Dirichlet CIFAR-10 with every
client in every round, and on Pets under the original protocol; with every sample kept, only a lower
test loss remains (Sect.~\ref{sec:robust}). We first describe what accompanies
the Pets gain under that protocol, where the mean violates nothing that could be repaired, and then
put both gains to the controls of Sect.~\ref{sec:controls}.

\subsection{The gain where the mean is feasible}\label{sec:pets}

On Pets the weighted mean harms no client to first order in any of the $\NPets$ round-seeds of
either trajectory (Table~\ref{tab:harmed}), so DualProj returns the mean there: recomputed on all
$\DualProjN$ stored Gramians of both arms, its weights depart from $p$ by at most $\DualProjDev$
relative, which is float round-off. The difference between the arms comes from UPGrad's projections
of individual updates, $\PairsPets\%$ of whose pairs conflict in a typical round.

\begin{figure}[!htb]
\centering
\includegraphics[width=\textwidth]{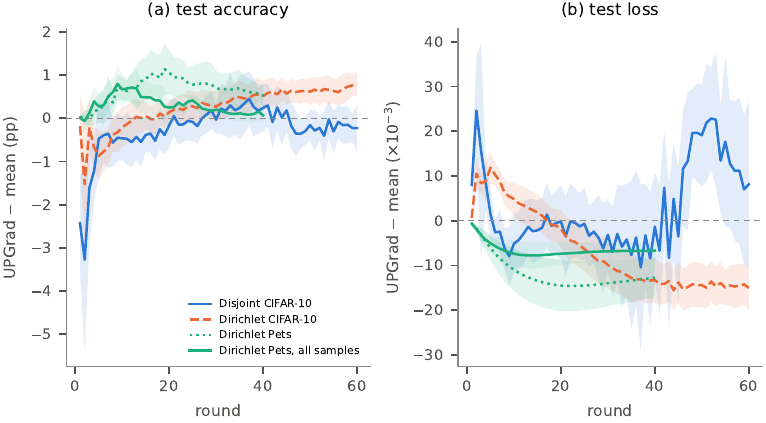}
\caption{Paired difference per round, UPGrad minus weighted mean, with $95\%$ confidence bands:
\textbf{(a)} centralized test accuracy, \textbf{(b)} centralized test loss. The solid green line
is Pets with every training sample kept (Sect.~\ref{sec:robust}).}
\label{fig:trajectory}
\end{figure}

\emph{On Pets, UPGrad's step is the better step for the training objective.} The centralized test loss is
lower under UPGrad after a single aggregation step from the common initialization
($\TrajPetsLossOne\times10^{-3}$, $\TrajPetsLossOneWins$ seeds, $\TrajPetsLossOneP$) and at every
round after it ($\TrajPetsLossLast\times10^{-3}$ at round~40, $\TrajPetsLossLastWins$); the
accuracy difference opens by round~10 ($\TrajPetsAccTen$~pp, $\TrajPetsAccTenWins$), peaks at
round~$\TrajPetsPeakRound$ ($\TrajPetsPeak$~pp) and narrows to $\AccPetsDelta$~pp
(Fig.~\ref{fig:trajectory}). At the same model $w_t$ on the UPGrad trajectory, UPGrad's step
lowers the clients' sample-weighted training loss ($L_k$ on the full training split, \texttt{eval}
mode) more than the equally long mean step in
$\StepPetsBetter$ of $\StepPetsN$ probed round-seeds. The gain on Pets is thus accompanied by an
advantage on the training objective at almost every probed step; whether that per-step advantage
accounts for the final accuracy difference, these measurements do not say. The CIFAR-10 regimes
behave differently: under Dirichlet partitioning UPGrad falls behind first ($\TrajDirAccFive$~pp at
round~5, $\TrajDirAccFiveWins$ seeds ahead), overtakes from round~$\TrajDirPosFrom$, and its step
is the better one in $\StepDirLastBetter$ seeds at round~60; on the disjoint split its step is
worse at round~1 ($\StepDisFirstBetter$) and the accuracy difference never stays positive.

\emph{UPGrad reweights the clients.} Its effective weights $u$ (Sect.~\ref{sec:operators}) move a
median $\WShiftPets\%$ of the weight mass on Pets and $\WShiftDir\%$ under Dirichlet CIFAR-10,
towards smaller clients (Spearman correlation between $u_k/p_k$ and $n_k$: $\WRhoSizePets$ and
$\WRhoSizeDir$). On the disjoint split, where shards are equal, it moves $\WShiftDis\%$, towards the
clients with the shorter updates.

Neither observation identifies a cause. A better step on the training objective could come from
the reweighting, from the particular direction UPGrad chooses, or from any perturbation of the
mean of that size; Sect.~\ref{sec:ctrlres} separates these readings with the controls of
Sect.~\ref{sec:controls}.

\subsection{Reweighting or orientation}\label{sec:ctrlres}

\begin{table}[!htb]
\caption{The two controls, seeds 1--10 paired with the main sweeps: Dirichlet CIFAR-10 ($K=20$,
$\alpha=0.5$) at round~60 and Pets under the original protocol (\texttt{drop\_last=True}) at
round~40. \emph{Rotation}: mean angle between the arm's direction and the
weighted mean of the transmitted updates. \emph{Mean infeasible}: share of round-seeds in which
that weighted mean would harm some client to first order, along the arm's own trajectory.
Differences from FedAvg in pp, with the seeds in which the arm is ahead and the paired
$t$-test $p$; worst-client accuracy resolves for no arm.}\label{tab:controls}
  \setlength{\tabcolsep}{4pt}
  \footnotesize
  \begin{tabular*}{\textwidth}{@{\extracolsep{\fill}}lcccc}
    \toprule
    & rotation & mean & \multicolumn{2}{c}{difference from FedAvg, pp (wins, $p$)} \\
    \cmidrule(lr){4-5}
    arm & ($^\circ$) & infeasible (\%) & centralized accuracy & client-mean accuracy \\
    \midrule
    \multicolumn{5}{l}{\emph{Dirichlet CIFAR-10}} \\
    FedAvg & 0 & 7.7 & & \\
    UPGrad & 16.2 & 84 & $+0.92$ (10/10, $<$0.001) & $+0.93$ (10/10, 0.009) \\
    uniform weights & 22.0 & 79 & $+0.96$ (9/10, 0.003) & $+1.13$ (9/10, 0.004) \\
    matched rotation & 9.9 & 22 & $+0.17$ (6/10, 0.337) & $+0.08$ (6/10, 0.598) \\
    \addlinespace[2pt]
    \multicolumn{5}{l}{\emph{Pets, original protocol}} \\
    FedAvg & 0 & 0 & & \\
    UPGrad & 6.3 & 0 & $+0.52$ (10/10, $<$0.001) & $+0.68$ (9/10, 0.007) \\
    uniform weights & 8.2 & 0 & $+1.37$ (10/10, $<$0.001) & $+1.36$ (10/10, $<$0.001) \\
    matched rotation & 4.5 & 0 & $-0.01$ (4/10, 0.786) & $-0.09$ (4/10, 0.340) \\
    \bottomrule
  \end{tabular*}

\end{table}

The controls ran under Dirichlet CIFAR-10 and on Pets under the original protocol, on seeds 1--10
paired with the main sweeps (same partition and initialization); on every stored Gramian the
matched rotation's angle equals UPGrad's to within $\CtrlPetsRotDev$ in cosine\suppref{Sect.~S3}.
In each configuration each control is compared with FedAvg and with UPGrad: four accuracy tests,
with a Bonferroni threshold of $\CtrlDirBonf$ (Table~\ref{tab:controls}).

\paragraph{Uniform weights reproduce the CIFAR-10 gain.} Weighting every client equally uses no
information about conflict, yet under Dirichlet CIFAR-10 it raises centralized accuracy by
$\CtrlDirUnif$~pp ($\CtrlDirUnifWins$ seeds, $\CtrlDirUnifP$), as much as UPGrad: UPGrad minus the
uniform control is $\CtrlDirUpVsUnif$~pp, equivalent to zero within $\pm0.5$~pp (TOST
$\CtrlDirUpVsUnifTOSTP$), and client-mean accuracy and test loss move with it. The two
reweightings have a similar size and sign: the smaller half of the clients holds $\WSmallDirP\%$ of
the weight under FedAvg, $\WSmallDirU\%$ under UPGrad and half under uniform weights. They also
lead to similar states. Along the uniform control's trajectory the weighted mean is infeasible in
$\CtrlDirUnifInf\%$ of round-seeds and UPGrad would turn it by $\CtrlDirWouldUnif^\circ$, against
$\CtrlDirUpInf\%$ and $\CtrlDirWouldUp^\circ$ along UPGrad's own and $\CtrlDirMeanInf\%$ and
$\CtrlDirWouldMean^\circ$ along FedAvg's, yet the uniform direction itself harms some client to
first order in only $\CtrlDirUnifHarm\%$ of round-seeds: in these states the infeasibility comes
with the sample-size weights.

\paragraph{A random rotation of matched angle does not.} On CIFAR-10 the matched rotation gains
$\CtrlDirRot$~pp ($\CtrlDirRotP$), and UPGrad is ahead of it by $\CtrlDirUpVsRot$~pp
($\CtrlDirUpVsRotWins$, $\CtrlDirUpVsRotP$), which passes the threshold. Its realized rotation,
$\CtrlDirRotRotation^\circ$ against UPGrad's $\CtrlDirUpRotation^\circ$, is smaller because the
angle is matched state by state and the control stays in states like FedAvg's; a rotation as large
as UPGrad's realized one was not run. The rotated direction harms some client to first order in
$\CtrlDirRotHarm\%$ of round-seeds, without a detectable change in accuracy or test loss.

\paragraph{On Pets, uniform weights do better than UPGrad.} Here the weighted mean is feasible in
every round-seed of every arm, and neither control's direction harms any client to first order.
Uniform weights raise centralized accuracy by $\CtrlPetsUnif$~pp ($\CtrlPetsUnifWins$,
$\CtrlPetsUnifP$), more than UPGrad's $\CtrlPetsUp$~pp (UPGrad minus uniform:
$\CtrlPetsUpVsUnif$~pp, $\CtrlPetsUpVsUnifP$, which passes the threshold), and lower the test loss
about twice as far ($\CtrlPetsUnifLoss\times10^{-3}$, against $\CtrlPetsUpLoss\times10^{-3}$).
UPGrad's weights follow client size only in part (Spearman $\WRhoSizePets$): the smaller half of
the clients holds $\WSmallPetsP\%$ of the weight under FedAvg, $\WSmallPetsU\%$ under UPGrad and
half under uniform weights. The matched rotation leaves accuracy unchanged ($\CtrlPetsRot$~pp,
$\CtrlPetsRotP$), raises the test loss in every seed and trails UPGrad by $\CtrlPetsUpVsRot$~pp
($\CtrlPetsUpVsRotWins$, $\CtrlPetsUpVsRotP$).

\paragraph{Reading.} In both configurations the gain needs neither attention to conflict nor
UPGrad's particular direction: weighting the clients equally achieves it, on Pets more of it, and
turning the mean by the angle UPGrad would apply, in a random direction, does not. This does not
show that UPGrad's gain \emph{is} its reweighting---two updates can reach the same accuracy by
different routes---but a reweighting blind to conflict suffices, so in these configurations the
gain is no evidence for the conflict rationale. The other CIFAR-10 results are consistent with this
reading without testing it: the disjoint split, whose equal shards already make the sample-size
weights uniform, shows no gain; the gain grows as $\alpha$ falls and client sizes become more
unequal; and it disappears with half the clients per round, where UPGrad moves about half as much
weight. On Pets the controls ran only under the original protocol; whether uniform weights keep
their gain with every sample kept, we did not test.

\section{Two exploratory configurations}\label{sec:exploratory}

Something in each configuration below---a step size, an averaging window, the configuration
itself---was fixed after seeing its runs; we report them as signals, not results.

\paragraph{One full-batch local step.} Replacing the local epoch with a single full-batch step makes
$g_k$ the gradient of $\widetilde L_k$ (relative norm error $\NormErrReg$ against
$\nabla L_k+\lambda w$, $\NormErrPlain$ against $\nabla L_k$). The constraint then binds almost
always: on the disjoint split the mean is infeasible in $\FBInfMain$ of $\FBNMain$ round-seeds at
$\eta_s=0.01$, against $\InfDisPct\%$ after an epoch---accumulating a local trajectory removes
most of the conflict before the server sees it. One step per round trains far less, so $\eta_s$ was
re-chosen by a one-seed scan as $0.5$ (Table~\ref{tab:fbscan}). Averaged over rounds 121--150,
because the runs oscillate at this rate, UPGrad does \emph{worse} on every headline metric in all
three seeds: centralized accuracy $\FBAccDelta$~pp ($\FBAccP$), worst-client accuracy
$\FBWorstDelta$~pp ($\FBWorstP$), spread $\FBStdDelta$~pp ($\FBStdP$). Where the first-order
guarantee is real, it did not buy a better outcome.

\paragraph{A configuration from FedGH's grid.} At $\alpha=0.1$, $E=5$, $K=10$ (3 seeds,
$\GHRounds$ rounds), where the mean is infeasible in $\GHInf$ of $\GHN$ round-seeds, the accuracy
metrics favour UPGrad in all three seeds (centralized $\GHAccDelta$~pp) and the fairness metrics
lean neither way; with $p$ between $\GHPmin$ and $\GHPmax$ across the seven metrics, three seeds
are a reason to power this configuration properly, not a result.

\section{Discussion}\label{sec:discussion}

\paragraph{What is established, and how firmly.} The mathematical statements
(Eq.~\eqref{eq:floor}, Examples~\ref{ex:dualproj} and~\ref{ex:meancos},
Propositions~\ref{prop:zero} and~\ref{prop:symmetric}) hold for any data; the measurements hold in
our configurations, with the uncertainty reported beside them. Neither accuracy gain survived every
check (Sect.~\ref{sec:robust}), and where the gains exist, a reweighting blind to conflict produces
them as well or better (Sect.~\ref{sec:ctrlres}): the outcome usually offered as evidence for
conflict-averse aggregation, a gain in accuracy, is here equally evidence for weighting small
clients more. Which route UPGrad's own gain takes, why the Pets gain depends on the data protocol,
and whether any of this holds beyond the configurations we varied, remains open.

\paragraph{The four quantities.} Collapsing them into ``is there conflict?'' lets an outcome in
either direction be credited to the presence or absence of conflict. They come apart within each
configuration (Sect.~\ref{sec:why}), move together along the Dirichlet concentration, which is why
that sweep attributes the gain to none of them, and part across configurations
(Sect.~\ref{sec:alpha}). The controls add that client weighting alone moves two of them: under
Dirichlet CIFAR-10, along the uniform control's trajectory the mean is infeasible about as often as
along UPGrad's and UPGrad would turn it about as far, so the trajectory dependence of feasibility
(Sect.~\ref{sec:binds}) does not require conflict-averse aggregation, only a different weighting.
We do not claim the quantities carry no information about one another---we fitted no
predictor---only that none determines the others or the outcome.

\paragraph{Using the diagnostics.} An evaluation of a conflict-averse aggregator in FL can report,
beside accuracy, six measurements. The first five are cheap beside the training they qualify, the
sixth costs two more runs per seed, and each changed a conclusion here: (i) conflict against the floor $-1/(K-1)$, with the share of conflicting pairs;
(ii) the feasibility of the mean along both trajectories, which here differ by an order of
magnitude; (iii) the alignment of the transmitted updates with the clients' gradients, under a
named objective and BatchNorm mode; (iv) the realized loss change at the candidate step rather
than the sign of a first-order margin; (v) the norm ratio, with the step norm equalized before
anything is attributed to direction, and the residual rotation with its Gramian counterfactuals;
and (vi) a reweighting control and a matched random rotation, which ask whether the direction
matters beyond the reweighting it implies and the size of the change it makes. The pipeline
deserves the same scrutiny as the operator (Sect.~\ref{sec:robust}).

\paragraph{Limitations.} The study is small: at most $20$ clients and $150$ rounds, two datasets, a
$62$\,k-parameter CNN trained from scratch and a partially fine-tuned ResNet-18. Heterogeneity is
varied only through the Dirichlet concentration on CIFAR-10 at $K=20$, which changes label skew,
client sizes and local step counts together and leaves feature skew untouched; client count and
local epochs are not varied independently, and one partial-participation rate was tried. The
controls ran in one configuration per dataset, on Pets only under the original protocol, and the
random rotation matches UPGrad's angle state by state, not its realized size. Each client visits
its data in the same order in every round, a property of our pipeline rather than of FedAvg; on
CIFAR-10 the $\DropShareDir\%$ of the data this order leaves out was never restored. We study one
non-conflicting operator, reimplement none of FedFV, FedGH or FedMGDA+, compare with no
drift-correction method and tune neither arm, so the paper answers question (A) only, and the seed
counts were not fixed in advance.

\section{Conclusion}\label{sec:conclusion}

Conflict-averse aggregation in FL is justified by a chain of four quantities that came apart in our
configurations: conflict between client updates did not determine whether their mean violated the
non-conflicting condition, the guarantee exact on the transmitted updates did not carry over to the
clients' gradients after a local epoch, and most of what UPGrad changed was the length of the
update. With only the direction differing, UPGrad's gains were small, confined to accuracy and
conditional: $\AccDirDelta$~pp under Dirichlet CIFAR-10 with every client in every round, and
$\AccPetsDelta$~pp on Oxford-IIIT Pet only under a protocol that never used a fifth of the training
data. Within CIFAR-10 the gain grew with heterogeneity, yet the split built for maximal conflict
showed none: the level of conflict did not predict the benefit. Weighting the clients equally
matched or exceeded both gains, while a random rotation of the same angle did neither. What carries
beyond these configurations is the protocol of Sect.~\ref{sec:discussion}: measure the four
quantities, at matched step norm and under named objectives, and compare with a reweighting that
ignores conflict before crediting an outcome to conflict. Whether effort is better spent on local
training, where the transmitted updates are formed, is a conjecture our data do not test.

\backmatter

\begin{appendices}

\section{Proofs and examples}\label{app:proofs}

\begin{example}[A feasible mean does not make UPGrad the identity]\label{ex:dualproj}
Let $g_1=(3,0)$ and $g_2=(-1,2)$ with equal weights. The mean $\gbar=(1,1)$ has margins $3$ and
$1$, so it is feasible and DualProj returns it. But $g_1^\top g_2=-3<0$, so neither row is in $C$:
$P_C(g_1)=(\tfrac{12}{5},\tfrac{6}{5})$, $P_C(g_2)=(0,2)$, and
$d_{\upgrad}=(\tfrac65,\tfrac85)\ne\gbar$.
\end{example}

\begin{example}[A positive mean cosine does not make UPGrad the identity]\label{ex:meancos}
Let $g_1=(1,0)$, $g_2=(0,1)$, $g_3=(2,-1)$ with equal weights. The mean pairwise cosine is
$1/(3\sqrt5)>0$ and $\gbar=(1,0)$, yet $P_C(g_2)=(\tfrac25,\tfrac45)$ and $P_C(g_3)=(2,0)$ give
$d_{\upgrad}=(\tfrac{17}{15},\tfrac{4}{15})$, a rotation of $13.2^\circ$ that norm equalization
does not remove.
\end{example}

\begin{proof}[Proof of Proposition~\ref{prop:zero}]
Take $L_1=\tfrac12(x+1)^2+y^2$, $L_2=\tfrac12(x-1)^2+y^2$ and $L_3=\tfrac12(y+1)^2$ at $w=(0,0)$,
with gradients $h_1=(1,0)$, $h_2=(-1,0)$, $h_3=(0,1)$. The dual cone is $C=\{d:\ d_x=0,\ d_y\ge0\}$.
UPGrad with equal weights projects $h_1$ and $h_2$ to $0$ and leaves $h_3$, so $d=(0,\tfrac13)$,
with margins $0$, $0$ and $\tfrac13$. For every $\eta>0$,
$L_1(w-\eta d)-L_1(w)=L_2(w-\eta d)-L_2(w)=\eta^2/9>0$.
\end{proof}

\begin{proposition}\label{prop:symmetric}
Let the rows have equal norms, equal weights and a common pairwise cosine $c$ with
$-1/(K-1)<c<0$. Then $d_{\upgrad}=\frac{1-c}{1+(K-2)c}\,\gbar$.
\end{proposition}

\begin{proof}
Take unit norms. We claim $P_C(g_i)=g_i+\beta\sum_{j\ne i}g_j$ with $\beta=-c/(1+(K-2)c)>0$. For
$j\ne i$, $g_j^\top P_C(g_i)=c+\beta\bigl(1+(K-2)c\bigr)=0$, and
$g_i^\top P_C(g_i)=1-(K-1)c^2/(1+(K-2)c)>0$ because $\bigl((K-1)c+1\bigr)(c-1)<0$ on the
interval; so the point is feasible. It differs from $g_i$ by a non-negative combination of the
normals $g_j$ of the constraints active at it, which are exactly those $j\ne i$; this is the KKT
condition of the projection, so the point is $P_C(g_i)$. Averaging over $i$ gives
$\bigl(1+(K-1)\beta\bigr)\gbar=\frac{1-c}{1+(K-2)c}\,\gbar$.
\end{proof}

\clearpage
\section{Additional tables}\label{app:tables}

\begin{table}[!htb]
\caption{All $\FamilySize$ paired tests: seven metrics in three regimes. $\Delta$ in pp (UPGrad
minus mean), W the seeds favouring UPGrad, $p$ the paired $t$-test and $p_{\rm sign}$ the exact
sign test. $\dagger$: $p<0.05$; $\ddagger$: below the Bonferroni threshold
$0.05/\FamilySize=\BonfThreshold$. Centralized, client-mean and sample-weighted accuracy are three
readings of one quantity, and worst client and worst decile of another, so the tests are not
independent; on the disjoint split, where shards are equal, client-mean and sample-weighted accuracy
coincide.}\label{tab:family}
  \setlength{\tabcolsep}{2pt}
  \fontsize{7}{8.5}\selectfont
  \begin{tabular*}{\textwidth}{@{\extracolsep{\fill}}lcccccccccccc}
    \toprule
    & \multicolumn{4}{c}{Disjoint} & \multicolumn{4}{c}{Dirichlet} & \multicolumn{4}{c}{Pets} \\
    \cmidrule(lr){2-5} \cmidrule(lr){6-9} \cmidrule(lr){10-13}
    Metric & $\Delta$ & W & $p$ & $p_{\rm sign}$ & $\Delta$ & W & $p$ & $p_{\rm sign}$ & $\Delta$ & W & $p$ & $p_{\rm sign}$ \\
    \midrule
    centralized       & $-0.23$ & 6/15 & 0.362 & 0.607 & $+0.78$ & 14/15 & $<$0.001$^{\ddagger}$ & $<$0.001 & $+0.52$ & 10/10 & $<$0.001$^{\ddagger}$ & 0.002 \\
    client mean       & $+0.003$ & 7/15 & 0.841 & 1.000 & $+0.88$ & 15/15 & $<$0.001$^{\ddagger}$ & $<$0.001 & $+0.68$ & 9/10 & 0.007$^{\dagger}$ & 0.021 \\
    sample-weighted   & $+0.003$ & 7/15 & 0.841 & 1.000 & $+0.77$ & 15/15 & $<$0.001$^{\ddagger}$ & $<$0.001 & $+0.63$ & 9/10 & 0.004$^{\dagger}$ & 0.021 \\
    worst client      & $+1.12$ & 9/15 & 0.302 & 0.607 & $+1.64$ & 12/15 & 0.047$^{\dagger}$ & 0.035 & $+1.18$ & 5/10 & 0.124 & 1.000 \\
    worst decile      & $+1.12$ & 9/15 & 0.302 & 0.607 & $+1.54$ & 11/15 & 0.038$^{\dagger}$ & 0.118 & $+0.73$ & 6/10 & 0.248 & 0.754 \\
    gap               & $-2.27$ & 9/15 & 0.174 & 0.607 & $-1.33$ & 13/15 & 0.125 & 0.007 & $-0.28$ & 5/10 & 0.820 & 1.000 \\
    spread      & $-0.98$ & 9/15 & 0.103 & 0.607 & $-0.37$ & 9/15 & 0.124 & 0.607 & $-0.04$ & 6/10 & 0.821 & 0.754 \\
    \bottomrule
  \end{tabular*}

\end{table}

\begin{table}[!htb]
\caption{The symmetric model of Eq.~\eqref{eq:symmetric} against every round with a negative mean
cosine: relative error of the predicted norm ratio. On Pets the mean cosine is never negative.}
\label{tab:symmetric}
  \setlength{\tabcolsep}{4pt}
  \footnotesize
  \begin{tabular*}{\textwidth}{@{\extracolsep{\fill}}lccccc}
    \toprule
    & & rounds with & \multicolumn{3}{c}{rel.\ error of predicted ratio} \\
    \cmidrule(lr){4-6}
    Regime & $K$ & $c<0$ & median & p90 & max \\
    \midrule
    Disjoint   & 5  & 897/900 & \textbf{2.1\%} & 4.7\% & 10.5\% \\
    Dirichlet  & 20 & 636/900 & 5.6\% & 16.5\% & 33.2\% \\
    Pets       & 20 & 0/400   & ---            & ---    & ---    \\
    \bottomrule
  \end{tabular*}

\end{table}

\begin{table}[!htb]
\caption{The two error sources of Eq.~\eqref{eq:split}, told apart by shrinking the step: median
over clients and probe rounds of the predicted loss change divided by the measured one ($1$ means
exact), predicted from the transmitted update or from the true gradient, with the interquartile
range of the former. In the one-step control the ratio from the update tends to
$1+\lambda w^\top d/h_k^\top d$, not $1$, because there $g_k=h_k+\lambda w$.}\label{tab:deltal}
  \setlength{\tabcolsep}{4pt}
  \footnotesize
  \begin{tabular*}{\textwidth}{@{\extracolsep{\fill}}llcccc}
    \toprule
    & prediction & \multicolumn{4}{c}{step, as a fraction of $\eta_s$} \\
    \cmidrule(lr){3-6}
    Configuration & uses & $1$ & $1/4$ & $1/16$ & $1/64$ \\
    \midrule
    Disjoint, 1 epoch            & from update & 502 & 652 & 644 & 644 \\
                                 & \quad interquartile & [-873, 1205] & [-336, 1901] & [-338, 1679] & [-339, 1676] \\
                                 & from gradient & 0.99 & 1.01 & 1.00 & 1.00 \\
    \addlinespace[2pt]
    Dirichlet, 1 epoch           & from update & 427 & 412 & 402 & 401 \\
                                 & \quad interquartile & [259, 792] & [235, 891] & [232, 857] & [233, 854] \\
                                 & from gradient & 1.03 & 1.01 & 1.00 & 1.00 \\
    \addlinespace[2pt]
    Disjoint, 1 full-batch step  & from update & 1.04 & 1.02 & 1.01 & 1.01 \\
                                 & \quad interquartile & [0.85, 1.15] & [1.00, 1.08] & [1.00, 1.04] & [1.00, 1.02] \\
                                 & from gradient & 1.03 & 1.01 & 1.00 & 1.00 \\
    \bottomrule
  \end{tabular*}

\end{table}

\begin{table}[!htb]
\caption{Scale and direction of the transmitted update (3 seeds). Per client: median norm ratio
and alignment. After aggregation: cosines between UPGrad's output on the updates $d(g)$, on the
true gradients $d(h)$ and on $s_k=(\lVert g_k\rVert/\lVert h_k\rVert)\,h_k$.}\label{tab:scale}
  \setlength{\tabcolsep}{4pt}
  \footnotesize
  \begin{tabular*}{\textwidth}{@{\extracolsep{\fill}}llcccrc}
    \toprule
    & & \multicolumn{2}{c}{per client} & \multicolumn{3}{c}{after aggregation} \\
    \cmidrule(lr){3-4}\cmidrule(lr){5-7}
    Regime & Rd & $\frac{\lVert g_k \rVert}{\lVert h_k \rVert}$ & $\cos(g_k,h_k)$ & $\cos\big(d(g), d(h)\big)$ & $\cos\big(d(s), d(h)\big)$ & $\cos\big(d(g), d(s)\big)$ \\
    \midrule
    Disjoint, $K=5$    & 1  & 414 & 0.65 & 0.094 & 1.0000 & 0.094 \\
                       & 10 & 45 & 0.71 & 0.077 & 0.9999 & 0.077 \\
                       & 30 & 34 & 0.57 & 0.018 & 0.9998 & 0.018 \\
                       & 60 & 41 & 0.40 & 0.032 & 0.9994 & 0.033 \\
    \midrule
    Dirichlet, $K=20$  & 1  & 315 & 0.85 & 0.249 & 0.9998 & 0.249 \\
                       & 10 & 43 & 0.45 & 0.251 & 0.9986 & 0.253 \\
                       & 30 & 49 & 0.37 & 0.124 & 0.9990 & 0.124 \\
                       & 60 & 68 & 0.28 & 0.079 & 0.9986 & 0.079 \\
    \bottomrule
  \end{tabular*}

\end{table}

\begin{table}[!htb]
\caption{Screening scan of the server rate for the one-step control: no-projection arm, disjoint
split, accuracy (\%) at three rounds and the peak over the first 20.}\label{tab:fbscan}
  \setlength{\tabcolsep}{4pt}
  \footnotesize
  \begin{tabular*}{\textwidth}{@{\extracolsep{\fill}}lccccll}
    \toprule
    & & \multicolumn{3}{c}{accuracy (\%) at round} & & \\
    \cmidrule(lr){3-5}
    $\eta_s$ & $n$ & 1 & 10 & 20 & peak & beyond the scan \\
    \midrule
    0.01  & 3 & 9.8 & 9.8 & 9.8 & 9.8 & $10.1$ at round $150$ \\
    0.1   & 1 & 9.4 & 10.0 & 10.7 & 10.7 & not run further \\
    0.5   & 1 & 9.6 & 12.3 & 19.0 & 21.2 & used for the comparison \\
    1.0   & 1 & 10.0 & 20.2 & 15.0 & 21.3 & unstable: peaks, then falls back \\
    \bottomrule
  \end{tabular*}

\end{table}

\FloatBarrier   
\end{appendices}

\bibliography{references}

\end{document}